\documentclass[letterpaper]{article} 
\usepackage[preprint]{aaai2027}
\usepackage[hyphens]{url}  
\usepackage{graphicx} 
\usepackage{natbib}  
\usepackage{caption} 
\usepackage{amsmath}
\usepackage{amsfonts}
\usepackage{booktabs}
\usepackage{multirow}
\usepackage{array}
\usepackage{xcolor}
\usepackage{color, colortbl}
\definecolor{mygray}{gray}{0.95}

\title{PAWN: Progress-Aware World Action Models for UAV Vision-and-Language Navigation}

\author{
    Hengxing Cai\textsuperscript{\rm 1},
    Jinhan Dong\textsuperscript{\rm 2}\equalcontrib,
    Changhao Nai\textsuperscript{\rm 3}\equalcontrib,
    Yijie Rao\textsuperscript{\rm 4},\\
    Jue Hou\textsuperscript{\rm 5},
    Zanyang Zhong\textsuperscript{\rm 1},
    Haidong Wang\textsuperscript{\rm 1},
    Renxin Zhong\textsuperscript{\rm 1}
}
\affiliations{
    \textsuperscript{\rm 1}School of Intelligent Systems Engineering, Sun Yat-Sen University, Shenzhen, China\\
    \textsuperscript{\rm 2}Beijing University Of Posts and Telecommunications, Beijing, China\\
    \textsuperscript{\rm 3}Harbin Institute of Technology\\
    \textsuperscript{\rm 4}Beihang University, Beijing, China\\
    \textsuperscript{\rm 5}Xiamen University
}

\begin{document}

\maketitle

\begin{abstract}

Unmanned Aerial Vehicle (UAV) Vision-and-Language Navigation (VLN) requires agents to autonomously complete navigation tasks in environments following natural language instructions.
Existing methods generate action sequences directly from current observations and instructions, lacking modeling of action execution consequences --- before execution, the model can neither anticipate changes in visual state nor evaluate the impact of different candidate sequences on navigation progress, thereby limiting further improvements in navigation performance.
To address this, we propose PAWN, a \textbf{P}rogress-\textbf{A}ware \textbf{W}orld Action Model (WAM) for UAV Vision-and-Language \textbf{N}avigation.
PAWN introduces a policy head and a world model head built on a shared vision-language backbone: the policy head generates multiple candidate action sequences, while the world model head conditions on each candidate to jointly predict future task progress and future visual latents after execution. At inference time, the candidate with the highest predicted progress is selected for execution, enabling the agent to make anticipatory decisions grounded in expected navigation outcomes before execution.
Experiments on AirNav and IndoorUAV demonstrate that PAWN achieves state-of-the-art performance among all compared methods, with ablation studies further validating the independent contribution of each component.
Real-world UAV experiments further demonstrate that PAWN can transfer from simulation to the real world without additional fine-tuning, validating its practical deployment feasibility.

\end{abstract}

\section{Introduction}

Unmanned Aerial Vehicle (UAV) Vision-and-Language Navigation (VLN) aims to enable UAV to autonomously complete navigation tasks in complex environments following natural language instructions.
This task has broad practical value across real-world applications, including rapid target localization in post-disaster rescue, automated inspection along described routes in infrastructure monitoring, and landmark-guided delivery in low-altitude logistics.

Existing UAV VLN methods, such as OpenFly~\cite{2025OpenFly} and AirVLN-R1~\cite{cai2026airnavlargescaleuavvisionandlanguage}, typically generate action sequences directly from current observations and language instructions, without modeling the consequences of action execution: when generating an action sequence, the model cannot anticipate how the visual state will change, nor foresee the navigation progress that will result from executing that sequence.
In scenarios where the target is temporarily invisible, severely occluded, or reachable only through long action sequences, such lack of foresight can lead to detours, oscillation, or deviation from the goal.

Meanwhile, embodied methods incorporating world models have been shown to facilitate stronger state representations and anticipatory decision-making in ground navigation, autonomous driving, and robot manipulation — as exemplified by Dreamer~\cite{hafner2019dream, hafner2023mastering}, GAIA-1~\cite{hu2023gaia}, and NWM~\cite{bar2025navigation}.
Within UAV VLN, however, this direction remains underexplored, with only a few recent concurrent works exploring this paradigm~\cite{zhao2026worldvln, zheng2026worldfly}.

Moreover, existing embodied world model methods typically employ future prediction only as auxiliary supervision during training, used to enhance visual or state representations; at inference time, action decisions still rely solely on the policy model's output, and the world model plays no role in evaluating candidate action sequences.
For UAV VLN, complex instructions often give rise to multiple seemingly plausible action sequences from the same observation.
These candidates may have similar generation probabilities, yet their execution outcomes can differ substantially — and current methods lack a mechanism to compare and select among them before execution.

To address these limitations, we propose PAWN, a \textbf{P}rogress-\textbf{A}ware \textbf{W}orld Action Model (WAM) for UAV Vision-and-Language \textbf{N}avigation, comprising a shared vision-language backbone, a policy head, and a world model head.
Given visual observations, language instructions, and the current pose state, the policy head generates candidate action sequences; the world model head conditions on each candidate to predict future task progress (quantifying navigation advancement) and future visual latents (constraining the model to learn visual state changes induced by the candidate actions).
At inference time, the policy head generates multiple candidate action sequences, the world model predicts future progress for each, and the system executes the candidate with the highest predicted progress.
Future prediction thus serves a dual role: providing anticipatory supervision during training and enabling candidate evaluation at inference time.

The main contributions of this paper are as follows:
\begin{itemize}
    \item We propose PAWN, a \textbf{P}rogress-\textbf{A}ware \textbf{W}orld Action Model for UAV Vision-and-Language \textbf{N}avigation that conditions on candidate action sequences to jointly predict future task progress and future visual latents, thereby modeling the potential consequences of action execution.
    \item We introduce a world-model-based candidate evaluation mechanism at inference time. Through multi-candidate generation and a future-progress selection mechanism, PAWN compares the expected navigation advancement of different candidates before execution.
    \item Experiments on AirNav and IndoorUAV demonstrate that PAWN achieves state-of-the-art performance across all evaluated methods and benchmark splits.
    \item Real-world UAV flight experiments validate PAWN's deployability in real environments, confirming that the method transfers from simulation to deployment without additional fine-tuning.
\end{itemize}

\section{Related Work}

\subsection{UAV Vision-and-Language Navigation}

Early UAV VLN followed the imitation learning paradigm from ground-based VLN, directly mapping visual observations and language instructions to action sequences, with representative works including Seq2Seq~\cite{anderson2018vision} and CMA~\cite{liu2023aerialvln}. 
These methods established foundational baselines for UAV VLN, but are limited by model capacity, resulting in limited generalization to complex instructions and unseen environments.

With the advancement of large language models (LLMs) and vision-language models (VLMs), UAV VLN has increasingly shifted toward large-scale pretrained model-based approaches, broadly divided into two categories.
The first adopts an end-to-end vision-language-action (VLA) paradigm, directly mapping visual observations and language instructions to low-level actions or waypoints; this includes VLA models transferred from general robot manipulation such as $\pi_0$~\cite{black2024pi_0} and OpenVLA~\cite{kim2025openvla}, as well as navigation-specific models that further incorporate explicit reasoning, such as Uni-NaVid~\cite{zhang2024uni} and AirVLN-R1~\cite{cai2026airnavlargescaleuavvisionandlanguage}.
The second organizes navigation in an agent-like, modular fashion, improving long-horizon instruction executability through explicit intermediate structures such as keyframe selection and sub-task decomposition, with representative works including OpenFly-Agent~\cite{2025OpenFly} and IndoorUAV-Agent~\cite{liu2025indooruavbenchmarkingvisionlanguageuav}, offering greater flexibility in open-vocabulary and semantically complex scenarios.

However, regardless of traditional imitation learning methods or recent VLN/VLA approaches, action generation predominantly follows a direct mapping paradigm of ``current observation + instruction $\rightarrow$ action'', without explicitly modeling the visual state changes and task progress resulting from action execution.
In scenarios where the target is temporarily invisible, severely occluded, or requires long action sequences, such lack of foresight easily leads to detours, oscillation, or deviation from the goal.
This limitation motivates us to incorporate a world model mechanism, enabling the agent to model the potential consequences of candidate action sequences before execution.

\subsection{World Models for Embodied Navigation}

The core idea of world models is to predict future states from past observations and actions, supporting richer spatiotemporal representation learning and anticipatory decision-making.
From the Dreamer series~\cite{hafner2019dream, hafner2023mastering}, which perform imagination-based planning in compact latent spaces, to generative world models that model future observations via video generation, this paradigm has demonstrated advantages for long-horizon decision-making in robot manipulation~\cite{yang2023unisim}, autonomous driving~\cite{hu2023gaia}, and ground navigation~\cite{bar2025navigation}. 

Within UAV VLN, the exploration of world models remains limited.
The most directly related works are WorldVLN~\cite{zhao2026worldvln} and WorldFly~\cite{zheng2026worldfly}. WorldVLN adapts a pretrained autoregressive latent video Transformer into a navigation model, predicting latent world state transitions from instructions and historical observations and decoding them into waypoints, with Action-aware Group Relative Policy Optimization (GRPO) applied for optimization.
WorldFly adopts a dual-branch coupled flow matching mechanism, jointly modeling video prediction and action generation within a unified VLA framework with a shared backbone.
These works provide initial evidence for the potential of world models in UAV VLN.

However, these works do not address the selection among candidate action sequences at inference time.
In fact, beyond representation learning, the future-state prediction capability of world models is naturally suited for evaluating candidate actions before execution.
The ``generate multiple candidates, then select via a evaluator'' paradigm has been explored in LLM reasoning~\cite{hao2023reasoning}, trajectory scoring for robot manipulation~\cite{du2024video}, and the imagine-and-rank usage of the navigation world model in ground navigation~\cite{bar2025navigation}; yet this capability has not been systematically exploited in UAV VLN.

The necessity of incorporating future prediction into inference-time candidate selection in UAV VLN manifests at two levels.
First, UAVs operate with greater action freedom than ground robots, and a single instruction often corresponds to multiple geometrically plausible flight trajectories; relying on a single policy sample risks being trapped in suboptimal solutions due to probability bias, while multi-candidate sampling before execution covers a broader decision space.
Second,  the relative quality of candidates typically only becomes apparent after several execution steps, making it difficult to assess from the current observation alone. World model predictions fill this gap by bringing execution outcomes forward to the decision stage, enabling the agent to compare the task progress of different candidates before execution, thereby alleviating difficulties caused by long-horizon error accumulation and temporary target invisibility.

These considerations motivate PAWN, which enables the world model to not only support representation learning during training, but also directly participate in candidate action sequence selection at inference time.

\section{Method}

\subsection{Problem Formulation}

At each timestep $t$, the agent receives an observation $O_t = (v_{1:t},\, L,\, S_t,\, A_{1:t-1})$ consisting of:
\begin{itemize}
    \item $v_{1:t} = (v_1, v_2, \ldots, v_t)$: a sequence of first-person RGB images captured from the start of the episode up to the current timestep;
    \item $L$: a natural language instruction describing the navigation goal;
    \item $S_t = (x_t, y_t, z_t, \theta_t)$: the current UAV state, comprising its three-dimensional coordinates and heading angle;
    \item $A_{1:t-1} = (A_1, A_2, \ldots, A_{t-1})$: the sequence of action sequences executed in previous timesteps.
\end{itemize}

The agent is required to learn a navigation policy $\pi$ that predicts a future action sequence $\hat{A}_t = \pi(O_t)$ based on the current observation.
The action space consists of a set of discrete navigation commands: \textbf{move forward}, \textbf{move backward}, \textbf{turn left}, \textbf{turn right}, and \textbf{stop}.
A navigation episode terminates when the agent issues the \textbf{stop} action or the maximum number of timesteps is reached.
The episode is considered successful if the UAV's final position is within 20 meters of the target location.

\subsection{Overview}

\begin{figure*}[t]
\centering
\includegraphics[width=\textwidth]{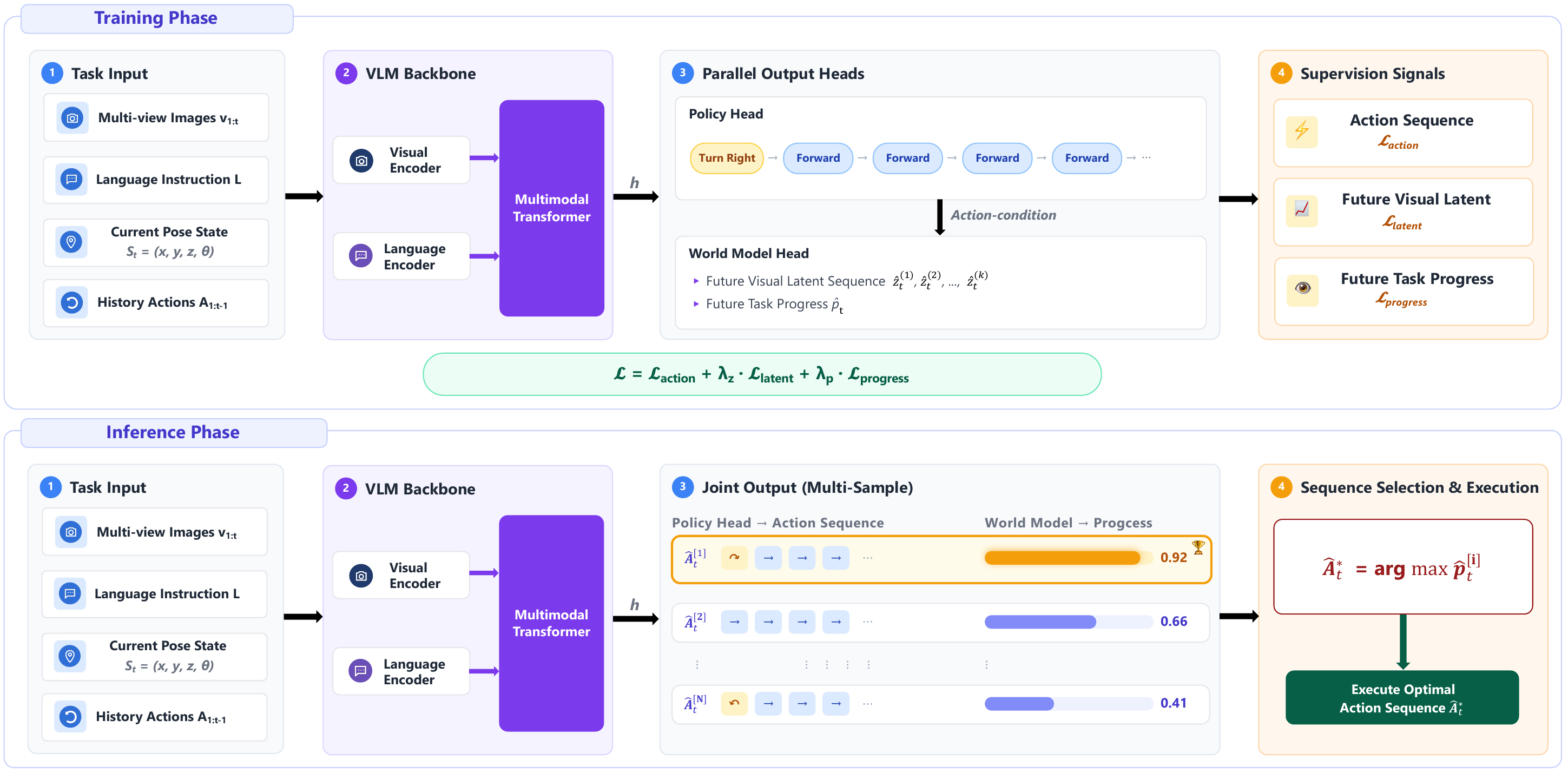}
\caption{Overview of the PAWN framework. The policy head generates candidate action sequences from the shared backbone representation, while the world model head predicts future visual latents and task progress conditioned on each candidate; at inference time, the candidate with the highest predicted progress is selected for execution.}
\label{fig:framework}
\end{figure*}

PAWN adopts the WAM paradigm, jointly modeling action generation and future state prediction.
As illustrated in Figure~\ref{fig:framework}, PAWN consists of three components: a shared visual-language backbone, a policy head, and a world model head.

At each timestep $t$, the shared backbone takes the observation $O_t$ as input and extracts a unified multimodal representation $\mathbf{h}_t$.
The policy head receives $\mathbf{h}_t$ and predicts a future action sequence $\hat{A}_t = \{ \hat{a}_t^{(1)}, \hat{a}_t^{(2)}, \dots, \hat{a}_t^{(k)} \}$ of length $k$, supervised by the expert trajectory via cross-entropy loss.
The world model head also takes $\mathbf{h}_t$ as input and, conditioned on $\hat{A}_t$, predicts two types of future states: (1) a sequence of future visual latents $\hat{\mathbf{z}}_t = \{ \hat{\mathbf{z}}_t^{(1)}, \hat{\mathbf{z}}_t^{(2)}, \dots, \hat{\mathbf{z}}_t^{(k)} \}$, serving as an auxiliary supervision signal that encourages the model to learn how actions alter the visual state; and (2) a future task progress score $\hat{p}_t$, quantifying how much the navigation task advances after executing the action sequence.
The detailed designs of these two prediction targets are described in Section~\ref{sec:world_model}.

At inference time, the policy head generates $N$ candidate action sequences from the same observation; the world model head predicts the future task progress for each candidate, and the system executes the sequence with the highest predicted progress.
The detailed inference procedure is described in Section~\ref{sec:inference}.


\subsection{Progress-Aware World Model}
\label{sec:world_model}

\subsubsection{Future Visual Latent Prediction}

PAWN introduces a future visual latent prediction task to constrain the world model to learn the latent visual state changes induced by action execution.
Given the current observation $O_t$ and the future action sequence $\hat{A}_t$, the world model predicts the sequence of visual latent representations $\hat{\mathbf{z}}_t$ corresponding to executing the action sequence.

Specifically, let $v_t^{(j)}$ denote the visual observation obtained after executing the first $j$ actions in the action sequence at timestep $t$.
We use a visual encoder to extract latent representations from these observations as supervision signals:
\begin{equation}
    \mathbf{z}_t^{(j)} = E_v\!\left(v_t^{(j)}\right), \quad j = 1, \ldots, k,
\end{equation}
where $E_v(\cdot)$ is the visual encoder. 

In the implementation, the future visual latent prediction head takes the representation of the current image from the shared backbone and the action sequence as input.
It models the relationship between the action sequence and the current visual state via cross-attention, and projects through a lightweight multi-layer perceptron (MLP), producing the predicted future visual latent sequence:
\begin{equation}
    \hat{\mathbf{z}}_t = \left\{ \hat{\mathbf{z}}_t^{(1)}, \ldots, \hat{\mathbf{z}}_t^{(k)} \right\}.
\end{equation}

During training, we construct a loss function based on the cosine similarity between predicted and ground-truth latents:
\begin{equation}
    \mathcal{L}_{\mathrm{latent}} = \frac{1}{k} \sum_{j=1}^{k} \left( 1 - \frac{\left(\hat{\mathbf{z}}_t^{(j)}\right)^\top \mathbf{z}_t^{(j)}}{\left\|\hat{\mathbf{z}}_t^{(j)}\right\|_2 \left\|\mathbf{z}_t^{(j)}\right\|_2} \right).
\end{equation}


\subsubsection{Future Task Progress Prediction}

Future task progress prediction estimates the degree of task completion after executing a given action sequence.
Given the current observation $O_t$ and the future action sequence $\hat{A}_t$, the world model predicts the future task progress $\hat{p}_t$ after executing the action sequence.

Let the expert action trajectory be
\begin{equation}
    A^* = \left\{ A^*_1, A^*_2, \ldots, A^*_M \right\},
\end{equation}
where $M$ denotes the total number of action sequences in the expert trajectory.
The task progress label $p_t$ is defined as the proportion of executed action sequences relative to the total expert trajectory length $M$, where $p_t$ specifically corresponds to the normalized completion ratio after executing $A^*_t$.
Compared to goal-distance supervision, this trajectory-normalized definition provides supervision at the action-chunk level, enabling the model to learn the association between action sequence quality and task advancement, rather than relying solely on spatial displacement to evaluate candidate sequences.


The progress prediction head takes the current multimodal context from the shared backbone and the action sequence as input, models the relationship between the action sequence and the current navigation context via cross-attention, then applies pooling and a lightweight MLP followed by sigmoid to produce the future task progress prediction $\hat{p}_t$.
During training, we use the true future task progress $p_t \in [0, 1]$ as a label, and supervise the progress prediction with binary cross-entropy:
\begin{equation}
    \mathcal{L}_{\mathrm{progress}} = -p_t \log \hat{p}_t - (1 - p_t) \log\!\left(1 - \hat{p}_t\right).
\end{equation}

Through this supervision objective, the world model learns the relationship among the current navigation context, the future action sequence, and the degree of task advancement.

\subsection{Training Objectives}

The overall training objective of PAWN is:
\begin{equation}
    \mathcal{L} = \mathcal{L}_{\mathrm{action}} + \lambda_z \mathcal{L}_{\mathrm{latent}} + \lambda_p \mathcal{L}_{\mathrm{progress}},
\end{equation}
where $\mathcal{L}_{\mathrm{action}}$ is the action sequence prediction loss of the policy head; $\mathcal{L}_{\mathrm{progress}}$ is the future task progress prediction loss; $\mathcal{L}_{\mathrm{latent}}$ is the future visual latent prediction loss; and $\lambda_p$, $\lambda_z$ are the corresponding loss weight coefficients.

\subsection{Inference Protocol}
\label{sec:inference}

At inference time, PAWN employs a multi-candidate sequence sampling and progress-guided selection mechanism.
Given the current observation $O_t$, the policy head generates $N$ candidate future action sequences via multinomial sampling (temperature $T=1.0$, $\text{top-}p=0.9$, $\text{top-}k=50$):
\begin{equation}
    \left\{ \hat{A}_t^{[i]} \right\}_{i=1}^{N}.
\end{equation}

For each candidate sequence $\hat{A}_t^{[i]}$, the world model predicts the future task progress after executing that sequence:
\begin{equation}
    \hat{p}_t^{[i]}, \quad i = 1, \ldots, N.
\end{equation}

PAWN then selects the candidate with the highest predicted progress:
\begin{align}
    i^* &= \arg\max_{i}\, \hat{p}_t^{[i]}, \\
    \hat{A}_t^{*} &= \hat{A}_t^{[i^*]}.
\end{align}

The agent executes the selected action sequence and, upon receiving a new observation, repeats the above procedure until the \textbf{stop} action is issued or the maximum number of timesteps is reached.

\section{Experiments}

\subsection{Experimental Setup}

\subsubsection{Datasets}

We evaluate PAWN on two UAV VLN benchmark datasets, covering outdoor and indoor navigation scenarios, respectively.
AirNav~\cite{cai2026airnavlargescaleuavvisionandlanguage} is a large-scale real-world outdoor UAV VLN dataset with natural and diverse navigation instructions.
IndoorUAV~\cite{liu2025indooruavbenchmarkingvisionlanguageuav} is an indoor UAV VLN benchmark built on the Habitat simulator, comprising over 1,000 scenes and 16,000 navigation trajectories, complementing AirNav in environment type.

\subsubsection{Baselines}

To comprehensively evaluate the effectiveness of PAWN, we compare it against three categories of baseline methods.
Traditional models include Random, Seq2Seq, and CMA.
Zero-shot VLMs include Qwen2.5-VL-7B~\cite{bai2025qwen25vltechnicalreport}, Qwen2.5-VL-32B~\cite{bai2025qwen25vltechnicalreport}, Qwen3-VL-235B-A22B~\cite{bai2025qwen3vltechnicalreport}, LLaMA-3.2-11B-Vision~\cite{grattafiori2024llama}, and GPT-5~\cite{singh2026openaigpt5card}, all of which perform navigation decisions directly under a zero-shot setting.
Fine-tuned VLN/VLA models include $\pi_0$~\cite{black2024pi_0}, Uni-NaVid~\cite{zhang2024uni}, OpenFly-Agent~\cite{2025OpenFly}, IndoorUAV-Agent~\cite{liu2025indooruavbenchmarkingvisionlanguageuav}, WorldFly~\cite{zheng2026worldfly}, and AirVLN-R1~\cite{cai2026airnavlargescaleuavvisionandlanguage}, all of which are fine-tuned on the training set before evaluation.
Detailed descriptions of all baselines are provided in Appendix~A.

\subsubsection{Evaluation Metrics}

We adopt four standard VLN evaluation metrics to assess model performance: Navigation Error (NE), Success Rate (SR), Oracle Success Rate (OSR), and Success weighted by Path Length (SPL).
Detailed definitions of these metrics are provided in Appendix~B.

\subsubsection{Implementation Details}

We use Qwen2.5-VL-7B as the visual-language backbone. All experiments are conducted on 8$\times$A100 GPUs, and each training epoch takes approximately 20 hours.
Key hyperparameter settings are provided in Appendix~C.

\subsection{Main Results}

\begin{table*}[ht]
\centering
\scriptsize
\setlength{\tabcolsep}{2pt}
\begin{tabular}{>{\centering\arraybackslash}p{4.3cm} | >{\centering\arraybackslash}p{0.77cm} >{\centering\arraybackslash}p{0.77cm} >{\centering\arraybackslash}p{0.77cm} >{\centering\arraybackslash}p{0.77cm} | >{\centering\arraybackslash}p{0.77cm} >{\centering\arraybackslash}p{0.77cm} >{\centering\arraybackslash}p{0.77cm} >{\centering\arraybackslash}p{0.77cm} | >{\centering\arraybackslash}p{0.77cm} >{\centering\arraybackslash}p{0.77cm} >{\centering\arraybackslash}p{0.77cm} >{\centering\arraybackslash}p{0.77cm}}
\toprule
\multirow{2}{*}{\textbf{Method}} & \multicolumn{4}{c|}{\textbf{Validation Seen}} & \multicolumn{4}{c|}{\textbf{Validation Unseen}} & \multicolumn{4}{c}{\textbf{Test Unseen}} \\
& NE$\downarrow$ & SR$\uparrow$ & OSR$\uparrow$ & SPL$\uparrow$ & NE$\downarrow$ & SR$\uparrow$ & OSR$\uparrow$ & SPL$\uparrow$ & NE$\downarrow$ & SR$\uparrow$ & OSR$\uparrow$ & SPL$\uparrow$ \\
\midrule
Random        & 221.96 & 0.76  & 5.68  & 0.67  & 223.06 & 0.76  & 4.73  & 0.68  & 217.25 & 0.77  & 5.57  & 0.69  \\
Seq2Seq$^\ast$       & 321.22 & 1.65  & 9.48  & 1.44  & 343.48 & 0.92  & 9.92  & 0.73  & 334.52 & 1.32  & 10.69 & 1.13  \\
CMA$^\ast$           & 184.68 & 5.21  & 15.84 & 4.80  & 201.12 & 3.88  & 16.07 & 3.47  & 189.43 & 4.63  & 17.48 & 4.16  \\
\midrule
Qwen2.5-VL-7B        & 182.06 & 1.96  & 2.36  & 1.82  & 192.49 & 1.59  & 1.76  & 1.41  & 184.24 & 1.70  & 1.95  & 1.53  \\
Qwen2.5-VL-32B       & 160.89 & 3.25  & 3.66  & 2.94  & 170.63 & 2.68  & 2.98  & 2.38  & 162.37 & 2.99  & 3.25  & 2.68  \\
Qwen3-VL-235B-A22B   & 157.59 & 5.58  & 9.23  & 5.18  & 168.47 & 5.14  & 8.42  & 4.63  & 155.56 & 5.21  & 8.44  & 4.75  \\
LLaMA-3.2-11B-Vision & 182.30 & 1.02  & 4.85  & 0.95  & 193.78 & 1.57  & 4.51  & 1.41  & 185.54 & 1.31  & 4.28  & 1.12  \\
GPT-5                & 150.51 & 3.10  & 3.46  & 2.80  & 156.07 & 2.48  & 2.60  & 2.16  & 152.48 & 2.76  & 2.93  & 2.49  \\
\midrule
$\pi_0^\ast$         & 141.23 & 6.13  & 14.42 & 5.49  & 162.92 & 4.71  & 13.88 & 4.15  & 146.66 & 5.02  & 13.62 & 4.41  \\
Uni-NaVid$^\ast$     & 112.25 & 20.56 & 38.02 & 18.45 & 129.25 & 16.60 & 33.45 & 14.71 & 131.20 & 15.89 & 34.25 & 14.18 \\
OpenFly-Agent$^\ast$ & 135.47 & 8.34  & 18.73 & 7.56  & 154.23 & 6.38  & 17.42 & 5.71  & 141.89 & 6.87  & 18.15 & 6.24  \\
IndoorUAV-Agent$^\ast$ & 167.87 & 3.72 & 3.87 & 3.32 & 178.67 & 2.70  & 2.88  & 2.32  & 169.67 & 3.66  & 3.88  & 3.28  \\
WorldFly$^\ast$      & 87.34  & 27.83 & 38.42 & 24.74 & 93.18  & 23.67 & 34.15 & 20.88 & 89.52  & 25.14 & 36.08 & 22.43 \\
AirVLN-R1$^\ast$     & 39.66  & 51.76 & 61.45 & 50.61 & 40.83  & 51.32 & 61.78 & 50.11 & 39.76  & 51.82 & 62.73 & 50.66 \\
\midrule
PAWN (Ours)$^\ast$   & \textbf{35.41} & \textbf{57.24} & \textbf{67.43} & \textbf{56.19} & \textbf{37.18} & \textbf{55.63} & \textbf{66.37} & \textbf{54.03} & \textbf{36.52} & \textbf{56.08} & \textbf{66.74} & \textbf{54.77} \\
\bottomrule
\end{tabular}
\caption{Comparison results on AirNav across three evaluation splits. $^\ast$ indicates models fine-tuned on the dataset.}
\label{tab:main_results}
\end{table*}

Table~\ref{tab:main_results} presents the comparison of PAWN against all baseline methods on AirNav.
Traditional models and zero-shot VLMs all achieve SR below 6\% on the test unseen split, indicating that neither simple sequence mapping nor general multimodal reasoning capabilities can achieve effective navigation performance without domain adaptation to UAV VLN tasks.

Among fine-tuned VLN/VLA models, performance varies significantly.
$\pi_0$, while capable of producing reasonable navigation behaviors, is primarily trained for ground robot scenarios, which limits its transferability.
In contrast, Uni-NaVid achieves stronger generalization through multi-task unified training, reaching an SR above 15\%; WorldFly, which integrates a dual-branch world model for joint video prediction and action generation during training, further improves to 25.14\% SR on test unseen split; AirVLN-R1 integrates supervised fine-tuning (SFT) and reinforcement fine-tuning (RFT) to introduce reasoning into navigation decisions, establishing the strongest baseline.

PAWN consistently outperforms all compared methods across all three evaluation splits.
On the test unseen split, PAWN achieves an SR of 56.08\% and an SPL of 54.77\%, validating its generalization to unseen environments.
Furthermore, PAWN also achieves competitive performance on the IndoorUAV indoor dataset, demonstrating strong cross-dataset generalization, with full IndoorUAV results provided in Appendix~D.

\subsection{Ablation Study and Hyperparameter Analysis}


\subsubsection{Component Ablation}

\begin{table}[t]
\centering
\small
\setlength{\tabcolsep}{4pt}
\begin{tabular}{lcccc}
\toprule
\textbf{Method} & NE$\downarrow$ & SR$\uparrow$ & OSR$\uparrow$ & SPL$\uparrow$ \\
\midrule
PAWN (full)              & \textbf{38.43} & \textbf{53.47} & \textbf{63.82} & \textbf{52.19} \\
\midrule
w/o World Model          & 47.49 & 40.20 & 53.21 & 39.15 \\
w/o Action Input         & 43.46 & 45.82 & 57.03 & 44.00 \\
w/o Future Latent        & 42.87 & 48.31 & 59.89 & 47.28 \\
w/o Progress Prediction  & 41.09 & 50.50 & 60.28 & 48.86 \\
\bottomrule
\end{tabular}
\caption{Component ablation on AirNav Test Unseen ($N=1$).}
\label{tab:ablation}
\end{table}

Table~\ref{tab:ablation} presents component ablation results under the $N=1$ setting, which isolates training-time contributions by excluding the inference-time candidate selection mechanism.
Removing the entire world model (w/o World Model) causes SR to drop from 53.47\% to 40.20\%, demonstrating that the world action model provides substantial auxiliary supervision that benefits policy learning.
Among the individual components, removing action input (w/o Action Input) leads to the largest single-component degradation (SR = 45.82\%): without conditioning on the candidate action sequence, the world action model cannot perceive the state differences induced by different action choices, losing its core conditional modeling capability.
Removing future visual latent prediction (w/o Future Latent) also causes notable degradation (SR = 48.31\%), indicating that explicitly supervising future visual states encourages the model to learn the correspondence between action sequences and visual state changes, thereby improving representation quality.
Removing progress prediction (w/o Progress Prediction) has a relatively small impact under $N=1$ (SR = 50.50\%), with its additional value further demonstrated through inference-time candidate selection in the next subsection.

\subsubsection{Effect of Candidate Sequence Number}
\label{sec:ablation_n}

\begin{figure}[t]
    \centering
    \includegraphics[width=\columnwidth]{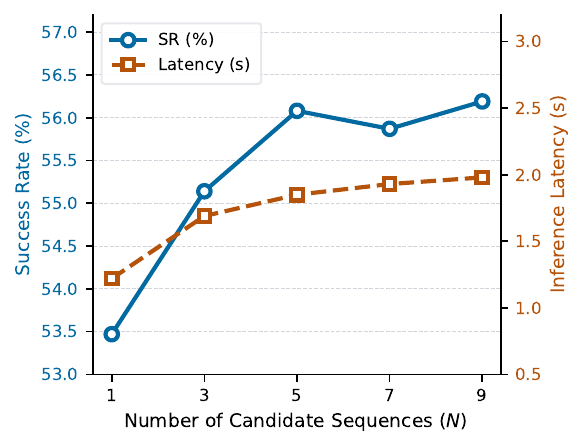}
    \caption{Effect of the number of candidate sequences $N$ on SR and inference latency. SR improves significantly up to $N=5$, then plateaus with minor fluctuations, while latency increases continuously.}
    \label{fig:n_analysis}
\end{figure}

Figure~\ref{fig:n_analysis} shows the effect of the number of candidate sequences $N$ on navigation performance and inference latency.
SR improves significantly from 53.47\% at $N=1$ to 56.08\% at $N=5$, demonstrating that multi-candidate progress-based selection effectively improves navigation decisions.
Beyond $N=5$, performance shows minor fluctuations ($N=7$: 55.87\%, $N=9$: 56.19\%), indicating that the selection benefit largely saturates.
Inference latency increases sublinearly with $N$: it rises notably from 1.22 s at $N=1$ to 1.69 s at $N=3$ as batch scheduling overhead is introduced, but the incremental cost diminishes rapidly thereafter (1.85 s at $N=5$, 1.93 s at $N=7$, 1.98 s at $N=9$).
This reflects the batch parallelism of the world action model, where visual encoding and prompt prefill are shared across all candidate passes and the marginal cost of additional candidates decreases as the batch grows.
We select $N=5$ as the default inference configuration, achieving a favorable balance between performance and efficiency.

\subsubsection{Loss Weight Sensitivity Analysis}

\begin{figure}[t]
    \centering
    \includegraphics[width=\columnwidth]{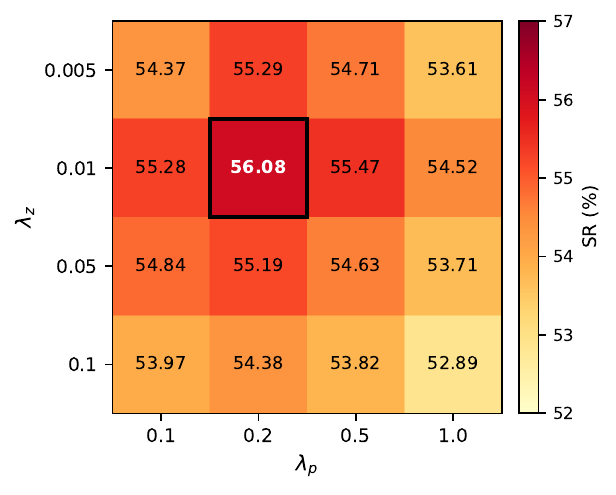}
    \caption{Sensitivity analysis of loss weights $\lambda_p$ and $\lambda_z$ on AirNav Test Unseen (SR\%, $N=5$).}
    \label{fig:lambda_heatmap}
\end{figure}

Figure~\ref{fig:lambda_heatmap} presents the sensitivity analysis of the progress prediction weight $\lambda_p$ and the visual latent prediction weight $\lambda_z$ in the training loss, conducted under the $N=5$ setting.
Across all 16 combinations with $\lambda_p \in \{0.1, 0.2, 0.5, 1\}$ and $\lambda_z \in \{0.005, 0.01, 0.05, 0.1\}$, SR ranges from 52.89\% to 56.08\%, with a maximum variation of 3.19 percentage points, demonstrating good robustness to hyperparameter choices.

\subsection{Qualitative Analysis}

\begin{figure}[t]
    \centering
    \includegraphics[width=\columnwidth]{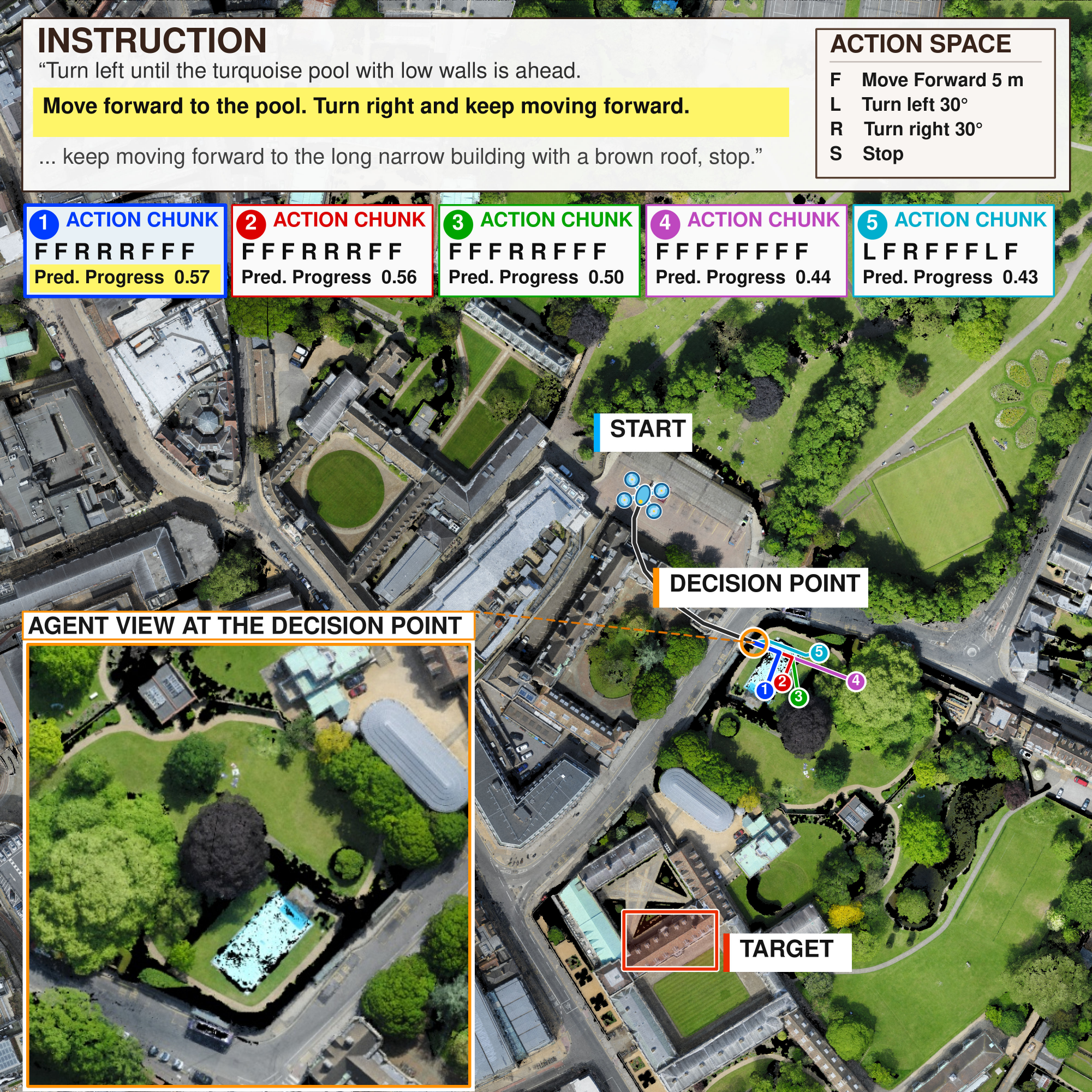}
    \caption{Qualitative analysis of PAWN's world action model on a representative AirNav scenario. Five candidate action sequences are generated at a decision point; predicted progress scores reveal a clear stratification based on turning timing, turn angle, and direction correctness.}
    \label{fig:qualitative}
\end{figure}

Figure~\ref{fig:qualitative} illustrates a representative scenario from AirNav, qualitatively demonstrating PAWN's ability to evaluate candidate action sequences.
The current navigation instruction requires the agent to fly forward toward the pool, then turn right and continue toward the target building (brown roof).
At the decision point shown, the policy head generates five candidate action sequences, and the world action model predicts future task progress for each, yielding a clear stratification:
Candidate 1, which executes a timely right turn and completes it fully, achieves the highest predicted progress (0.57);
Candidate 2, which turns slightly later but also completes the full turn, shows a marginal drop (0.56);
Candidate 3, with similar turn timing but insufficient turn angle, scores further lower (0.50);
Candidate 4, which proceeds straight ahead and misses the turn entirely, drops notably (0.44);
Candidate 5, which turns in the wrong direction, scores lowest (0.43).
These results demonstrate that PAWN can not only distinguish the correctness of action directions, but also perceive differences in turning timing and turn angle, effectively guiding the system to select the optimal candidate sequence before execution.

\subsection{Real-World UAV Experiments}

\subsubsection{Setup}

We conduct real-world UAV flight experiments with PAWN in outdoor environments without any additional fine-tuning, comprising 200 navigation tasks that cover a variety of path lengths and target object types (success threshold: 20\,m, consistent with the simulation benchmark).

The UAV platform is a DJI Matrice 350 RTK equipped with a WIT-Motion SPL06 barometric altimeter, a Beitian BT-560 real-time kinematic positioning module, and a Raspberry Pi Camera Module 3 monocular RGB camera. The communication link supports dual-mode cellular and LoRa redundancy. Model inference is deployed on a cloud server.

The system operates in a three-stage pipeline: (1) \textbf{Perception}: at each timestep, the UAV captures a first-person RGB image and records its current position and heading; (2) \textbf{Inference}: the perception data is transmitted to the cloud server via cellular network, where the server integrates the current observation with historical context and invokes the model for inference; (3) \textbf{Action Execution}: the server sends the predicted action commands to the UAV flight controller, which drives the UAV to execute the corresponding actions.

\subsubsection{Results}

\begin{table}[t]
\centering
\resizebox{\columnwidth}{!}{%
\begin{tabular}{lcccc}
\toprule
\textbf{Method} & NE$\downarrow$ & SR$\uparrow$ & Memory (GB) & Latency (s/step) \\
\midrule
Seq2Seq            & 205.3 & 2/200  & 0.18  & 0.018 \\
CMA                & 189.6 & 4/200  & 0.22  & 0.021 \\
Qwen2.5-VL-7B      & 151.4 & 10/200 & 16.20 & 1.13  \\
Qwen2.5-VL-32B     & 132.7 & 18/200 & 62.95 & 4.92 \\
GPT-5              & 121.3 & 22/200 & /     & 2.07     \\
\midrule
PAWN (Ours)        & 89.2  & 54/200 & 16.92 & 1.85  \\
\bottomrule
\end{tabular}%
}
\caption{Comparison of navigation performance, deployment cost, and inference efficiency in real-world UAV experiments.}
\label{tab:realworld}
\end{table}

Table~\ref{tab:realworld} presents a comprehensive comparison of all methods across navigation performance, deployment cost, and inference efficiency.

\noindent\textbf{Navigation Performance.}
PAWN achieves a success rate of 54/200 with a navigation error of 89.2\,m, significantly outperforming traditional imitation learning methods (Seq2Seq: 2/200, CMA: 4/200) and zero-shot VLM baselines (Qwen2.5-VL-7B: 10/200, Qwen2.5-VL-32B: 18/200, GPT-5: 22/200), achieving the best results among all methods evaluated in real-world deployment.

\noindent\textbf{Deployment Cost.}
PAWN runs locally on a single RTX 4090 with 16.92\,GB of GPU memory, without relying on cloud APIs, facilitating compliance with system controllability and data privacy requirements.
In contrast, Qwen2.5-VL-32B requires approximately 62.95\,GB of GPU memory, and GPT-5 relies on cloud APIs and does not support local deployment.

\noindent\textbf{Inference Efficiency.}
At $N=5$, PAWN's inference latency is 1.85\,s/step; despite performing parallel forward passes over $N$ candidate sequences, the latency remains lower than GPT-5, which is additionally subject to network fluctuations, and falls within an acceptable range.

\section{Conclusion}

This paper presents PAWN, a progress-aware WAM framework for UAV VLN.
To address the lack of modeling of action consequences in existing methods, PAWN introduces a world model head on top of a shared vision-language backbone, jointly predicting future visual latents and future task progress to provide anticipatory supervision for policy learning during training.
At inference time, PAWN employs a multi-candidate generation and progress-based selection mechanism, enabling the agent to compare the potential outcomes of candidate sequences before execution and make anticipatory navigation decisions.
Experiments on two benchmarks, AirNav and IndoorUAV, demonstrate that PAWN achieves state-of-the-art performance among all compared methods, and ablation studies further validate the independent contribution of each component.
Real-world experiments additionally show that PAWN can transfer from simulation to the real world without additional fine-tuning, demonstrating practical deployment feasibility.

\section{Limitation}

\noindent\textbf{Inference Resource Requirement.}
At inference time, PAWN performs parallel forward passes over $N$ candidate action sequences.
When $N$ is large, the computational resources required for batch inference may exceed the capacity of edge devices, limiting practical deployment in resource-constrained scenarios.

\noindent\textbf{Progress Signal Limitations.}
The task progress label is defined as the proportion of completed action sequences relative to the expert trajectory — a sequence-level proxy that does not directly measure spatial distance to the goal, and may not accurately reflect true task advancement when instructions are ambiguous or expert trajectories contain detours.
Furthermore, since the progress predictor is trained solely on expert trajectories, its scoring accuracy on candidate sequences that deviate substantially from expert behavior at inference time remains to be validated.

\noindent\textbf{Simulation-to-Real Gap.}
PAWN is trained on a simulated dataset and deployed directly on a real UAV without additional fine-tuning.
Discrepancies between simulation and the real world — including lighting variation, camera shake, and GPS drift — are not explicitly addressed, which may limit performance in real-world deployment.

\bibliography{aaai2026}

\clearpage
\appendix
\section{Baseline Method Details}
\label{appendix:baselines}

\textbf{Traditional Models.}
\begin{itemize}
    \item \textbf{Random} samples actions uniformly at random from the action space at each timestep, serving as a lower-bound reference without using any visual or language information.
    \item \textbf{Seq2Seq}~\cite{anderson2018vision} is a classic sequence-to-sequence navigation model that directly maps visual observation-instruction pairs to action sequences.
    \item \textbf{CMA}~\cite{liu2023aerialvln} is a Cross-Modal Alignment baseline designed for VLN, effectively fusing language and visual information.
\end{itemize}

\noindent\textbf{Zero-shot Models.}
\begin{itemize}
    \item \textbf{Qwen2.5-VL-7B/32B}~\cite{bai2025qwen25vltechnicalreport} are open-source vision-language model series released by Alibaba, supporting multimodal image-text understanding and reasoning.
    \item \textbf{Qwen3-VL-235B-A22B}~\cite{bai2025qwen3vltechnicalreport} is a large-scale mixture-of-experts (MoE) vision-language model released by Alibaba, showing significant improvements in visual reasoning and instruction following over smaller models.
    \item \textbf{LLaMA-3.2-11B-Vision}~\cite{grattafiori2024llama} is an open-source multimodal large language model released by Meta, capable of joint image-text understanding.
    \item \textbf{GPT-5}~\cite{singh2026openaigpt5card} is OpenAI's latest large language model, with stronger multimodal reasoning capabilities.
\end{itemize}

\noindent\textbf{Fine-tuned VLN/VLA Models.}
\begin{itemize}
    \item $\boldsymbol{\pi_0}$~\cite{black2024pi_0} is a general-purpose robotic VLA model proposed by Physical Intelligence, combining a pretrained vision-language model backbone with a flow-matching policy head for continuous action prediction.
    \item \textbf{Uni-NaVid}~\cite{zhang2024uni} is a unified video-foundation VLA navigation model that consolidates different navigation sub-tasks, including instruction following, object search, visual question answering, and human following, into a single input-output format.
    \item \textbf{OpenFly-Agent}~\cite{2025OpenFly} is a keyframe-aware aerial navigation model proposed by the OpenFly platform, identifying and prioritizing key visual frames during flight to assist navigation decisions.
    \item \textbf{IndoorUAV-Agent}~\cite{liu2025indooruavbenchmarkingvisionlanguageuav} is the baseline method proposed alongside the IndoorUAV benchmark, decomposing complex navigation instructions into multiple sub-tasks executed in stages, evaluated in a continuous indoor UAV environment built on the Habitat simulator.
    \item \textbf{WorldFly}~\cite{zheng2026worldfly} is a world-model-based VLA framework for UAV navigation that employs a dual-branch coupled architecture to jointly predict future visual states and navigation actions via flow matching, enabling anticipatory decision-making through future scene imagination.
    \item \textbf{AirVLN-R1}~\cite{cai2026airnavlargescaleuavvisionandlanguage} is a UAV VLN model trained via SFT and RFT, introducing reasoning capability into the navigation decision process.
\end{itemize}

\section{Evaluation Metrics}
\label{appendix:metrics}

\textbf{NE (Navigation Error)}$\downarrow$: the distance between the UAV's final stopping position and the target position. A lower value indicates that the UAV's stopping position is closer to the target.

\noindent\textbf{SR (Success Rate)}$\uparrow$: the percentage of navigation episodes in which the UAV's final position is within a threshold distance of the target position.

\noindent\textbf{OSR (Oracle Success Rate)}$\uparrow$: the percentage of episodes in which the UAV comes within a threshold distance of the target at any point during the navigation trajectory, regardless of where it finally stops. OSR measures whether the UAV ever got close to the target, without considering whether it stopped correctly.

\noindent\textbf{SPL (Success weighted by Path Length)}$\uparrow$: success rate weighted by path efficiency, defined as the success-weighted average of the ratio between the shortest path length and the actual path length taken. SPL jointly measures navigation accuracy and efficiency.

\noindent\textbf{NDTW (Normalized Dynamic Time Warping)}$\uparrow$: trajectory fidelity metric that computes the dynamic time warping distance between the agent's path and the expert reference path, normalized to $[0, 1]$. Higher values indicate greater alignment with the reference trajectory.

\section{Implementation Details}
\label{appendix:implementation}

\begin{table}[h]
\centering
\begin{tabular}{ll}
\toprule
Hyperparameter & Value \\
\midrule
Optimizer & AdamW \\
Learning rate & 1e-4 \\
Batch size & 24 \\
Epochs & 2 \\
Action sequence length $k$ & 8 \\
Loss weight $\lambda_p$ & 0.2 \\
Loss weight $\lambda_z$ & 0.01 \\
\bottomrule
\end{tabular}
\caption{Key hyperparameters of PAWN.}
\label{tab:hyperparameters}
\end{table}


\onecolumn
\section{Cross-Dataset Generalization: Results on IndoorUAV}
\label{appendix:indooruav}

\begin{table}[!ht]
\centering
\scriptsize
\setlength{\tabcolsep}{2pt}
\begin{tabular}{>{\centering\arraybackslash}p{3.8cm} | >{\centering\arraybackslash}p{0.77cm} >{\centering\arraybackslash}p{0.77cm} >{\centering\arraybackslash}p{0.77cm} >{\centering\arraybackslash}p{0.92cm} | >{\centering\arraybackslash}p{0.77cm} >{\centering\arraybackslash}p{0.77cm} >{\centering\arraybackslash}p{0.77cm} >{\centering\arraybackslash}p{0.92cm}}
\toprule
\multirow{2}{*}{\textbf{Method}} & \multicolumn{4}{c|}{\textbf{Test Seen}} & \multicolumn{4}{c}{\textbf{Test Unseen}} \\
& NE$\downarrow$ & SR$\uparrow$ & OSR$\uparrow$ & NDTW$\uparrow$ & NE$\downarrow$ & SR$\uparrow$ & OSR$\uparrow$ & NDTW$\uparrow$ \\
\midrule
Seq2Seq$^\ast$          & 10.18 & 0.89  & 2.23  & 1.41  & 11.60 & 0.41  & 1.85  & 1.23  \\
CMA$^\ast$              & 11.60 & 1.56  & 8.48  & 1.39  & 12.15 & 1.64  & 10.47 & 0.81  \\
\midrule
Qwen2.5-VL-7B           & 7.31  & 1.02  & 1.53  & 5.07  & 7.40  & 0.00  & 0.00  & 4.75  \\
GPT-5                   & 6.90 & 2.31  & 4.23 & 9.54 & 7.17  & 2.25 & 2.25 & 9.3 \\
\midrule
$\pi_0^\ast$            & 8.35  & 2.92  & 11.50 & 11.87 & 8.81  & 2.83  & 10.02 & 11.69 \\
Uni-NaVid$^\ast$        & 7.38  & 6.71  & 13.24 & 13.08 & 8.29  & 4.85  & 12.17 & 11.92 \\
OpenFly-Agent$^\ast$    & 8.17  & 4.12  & 10.96 & 10.63 & 8.83  & 2.58  & 9.45  & 10.14 \\
IndoorUAV-Agent$^\ast$  & 6.62  & 7.29  & 12.83 & 17.19 & 7.27  & 5.06  & 13.49 & 15.65 \\
WorldFly$^\ast$         & 6.47  & 13.28 & 21.84 & 16.84 & 6.83  & 12.41 & 20.57 & 15.73 \\
AirVLN-R1$^\ast$        & 5.80  & 19.23 & 29.62 & 25.91 & 5.58  & 20.22 & 29.59 & 27.15 \\
\midrule
PAWN (Ours)$^\ast$      & \textbf{5.43} & \textbf{22.87} & \textbf{33.14} & \textbf{28.96} & \textbf{5.19} & \textbf{23.91} & \textbf{33.52} & \textbf{30.82} \\
\bottomrule
\end{tabular}
\caption{Comparison results on IndoorUAV-VLN with representative baselines. $^\ast$ indicates models fine-tuned on the dataset.}
\label{tab:indooruav}
\end{table}

\end{document}